\documentclass[runningheads]{llncs}
\usepackage{graphicx}
\usepackage{cite}
\usepackage{algorithm}
\usepackage{booktabs}
\usepackage{multirow}
\usepackage{array}
\usepackage{subfigure}
\usepackage{algpseudocode}
\usepackage{amsmath}
\usepackage{graphics}
\usepackage{epsfig}
\usepackage{color}
\usepackage{bm}
\usepackage{amsmath}
\usepackage{multirow}
\usepackage{graphicx}
\usepackage{array}
\usepackage{subfigure}
\usepackage{algpseudocode}
\usepackage{amsmath}
\usepackage{graphics}
\usepackage{epsfig}
\usepackage[colorlinks,
linkcolor=blue,
anchorcolor=black,
citecolor=black]{hyperref}
\usepackage{color}
\usepackage{bm}
\usepackage{amsfonts}
\usepackage{amsmath}

\usepackage{graphicx}
\usepackage[outdir=./]{epstopdf}
\usepackage{authblk}
\usepackage{threeparttable}

\makeatletter
\newenvironment{breakablealgorithm}
{
	\begin{center}
		\refstepcounter{algorithm}
		\hrule height.8pt depth0pt \kern2pt
		\renewcommand{\caption}[2][\relax]{
			{\raggedright\textbf{\ALG@name~\thealgorithm} ##2\par}%
			\ifx\relax##1\relax 
			\addcontentsline{loa}{algorithm}{\protect\numberline{\thealgorithm}##2}%
			\else 
			\addcontentsline{loa}{algorithm}{\protect\numberline{\thealgorithm}##1}%
			\fi
			\kern2pt\hrule\kern2pt
		}
	}{
		\kern2pt\hrule\relax
	\end{center}
}
\makeatother
\begin{document}
	\title{More but Correct: Generating Diversified and Entity-revised Medical Response\protect \footnote[1]{This work is supported by the National Key Research and Development Pro-ject(2018YFB1305200), the National Natural Science Fund of China(61801178) and the project of  Hunan Provincial Health Commission(202114010841).}\vspace{-0.5cm}}
	%
	%
	\author{Bin Li\inst{1}  \and
		Encheng Chen\inst{2}  \and
		Hongru Liu\inst{3}  \and
		Yixuan Weng \protect \footnote[6]{Email: wengsyx@qq.com.} \and
		Bin Sun\inst{1} \and
		Shutao Li\inst{1}\protect \footnote[4]{Corresponding author.} \and
		Yongping Bai\inst{4} \and
		Meiling Hu\inst{5}  
	\vspace{-0.1cm}}
	\renewcommand{\lastandname}{\unskip,}
	\authorrunning{Li et al.}
	\titlerunning{Generating Diversified and Entity-revised Medical Response}
	%
	\institute{College of Electrical and Information Engineering, Hunan University \\
		\email{\{libincn,shutao\_li,sunbin611\}@hnu.edu.cn} \and
		School of Mathematics, Sun Yat$-$Sen University \\
		\email{1510605163@qq.com} \and JD Technology\\
		\email{liuhongru3@jd.com}\and
		Xiangya Hospital of Central South University \\
		\email{baiyongping@csu.edu.cn} \\ \and
		Teaching and Research Section of Clinical Nursing, Xiangya Hospital of Centrail South University \\		\email{humeiling0704@126.com}	
	}
	
	\maketitle              
	\vspace{-0.5cm}
	\begin{abstract}
	Medical Dialogue Generation (MDG) is intended to build a medical dialogue system for intelligent consultation, which can communicate with patients in real-time, thereby improving the efficiency of clinical diagnosis with broad application prospects.
	This paper presents our proposed framework for the Chinese MDG organized by the 2021 China conference on knowledge graph and semantic computing (CCKS) competition, which requires generating context-consistent and medically meaningful responses conditioned on the dialogue history. 
	In our framework, we propose a pipeline system composed of entity prediction and entity-aware dialogue generation, by adding predicted entities to the dialogue model with a fusion mechanism, thereby utilizing information from different sources. At the decoding stage, we propose a new decoding mechanism named Entity-revised Diverse Beam Search (EDBS) to improve entity correctness and promote the length and quality of the final response.
	The proposed method wins both the CCKS and the International Conference on Learning Representations (ICLR) 2021 Workshop Machine Learning for Preventing and Combating Pandemics (MLPCP) Track 1 Entity-aware MED competitions, which demonstrate the practicality and effectiveness of our method.	
	\vspace{-0.2cm}
	\keywords{Medical Entity Prediction \and Entity-aware Fusion Dialogue Generation \and Entity-revised Diverse Beam Search.}
	\end{abstract}
	%
	%
	%

	\section{Introduction}
	\vspace{-0.2cm}
	During the COVID-19 epidemic, problems such as shortage of medical resources, hard burdens in doctors, and long waiting time for patients have existed in China. As a result, building an automatic response medical dialogue system is beneficial to improving the efficiency of clinical consultation and reducing the burden on doctors. To promote the research of Chinese medical dialogue generation, the 15th China Conference on Knowledge Graph and Semantic Computing (CCKS 2021) sets Task 11 for Entity-containing Medical Dialogue Generation, where participants are required to build the dialogue generation model based on the doctor-patient dialogue corpus in Gastroenterology.\par
	In recent years, medical dialogue generation has attracted more and more attention, due to its wide range of applications\cite{1,2,3,4,5}. However, to realize the real application landing, that is, to make the model imitate the  real doctor, there are two problems that need to be solved urgently. One is that the model needs to be able to give a reasonable response,  which often involves correct medical entity information\cite{1,2}. The other is that the model needs to imitate human thinking habits to generate responses, which are often long\cite{3,4}. To achieve this, we propose a pipeline system that contains two parts, including entity prediction and entity-aware dialogue response generation. Specifically, our contributions in this work can be summarized as follows
	\begin{itemize}
	\item 
	We build a framework for medical dialogue generation, adopting the pipeline structure with strong flexibility to get high-quality responses.
	\item
	An Encoding fusion module is developed for adaptively controlling the encodings of different sources, making full use of the medical entity information in the dialogue.
	\item
	We proposed Entity-revised Diversity Beam Search (EDBS), which can improve the diversity of final responses, while keeping the complete predicted entity information.
	\end{itemize}
	\vspace{-0.5cm}
	\section{Related Work}
	\vspace{-0.15cm}
	Medical dialogue generation has made great progress in recent years. Early research mainly focuses on task-oriented dialogue systems\cite{1,2}, which emphasize automatic disease diagnosis in high accuracy. However, the final response is often in the template, which requires vast human labor in designing. Subsequently, many studies begin to explore automatic response. Zeng et al.\cite{3} develop a dialogue system for COVID-19. Liu et al. \cite{4} construct the dialogue system in the field of Gastrointestinal. Nonetheless, it is not easy for the model to imitate the doctor. The simple Seq2Seq\cite{6} structure cannot make good use of the knowledge information with reasoned entities. Besides, it is easy to generate shorter responses based on Beam Search \cite{7} during the decoding, while the predefined conditions will be shifting with Diverse Beam Search (DBS)\cite{8}. Different from the previous work, we improve the original Seq2Seq dialogue generation model with the encoding fusion mechanism. Specifically, we add contextual encodings and predicted entities encoding to the dialogue generation with encoding fusion, thereby making full use of information from different sources. At the same time, we propose a decoding mechanism Entity-revised Diverse Beam Search (EDBS) to improve the overall quality of the final response in terms of F1 and BELU scores.
	\begin{figure}[t]
	\centering
	\includegraphics[scale = 0.45]{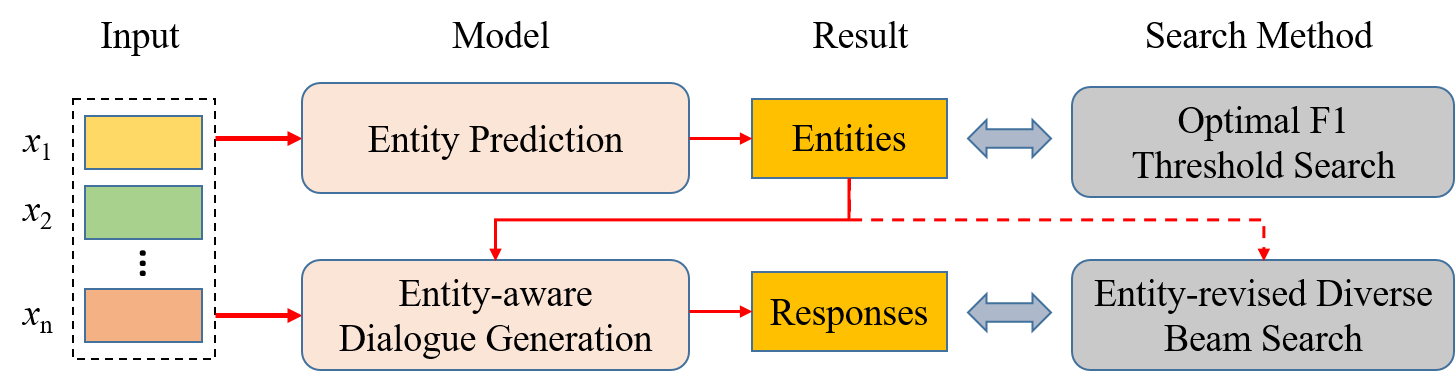}
	\caption{Overview of System Framework.} 
	\label{frame1}
	\end{figure}
	\vspace{-1.2cm}
	\section{Method}
	In this section, we will first illustrate the framework of our method, then we depict each component of the framework in detail. Finally, we present the fusion strategy for improvement.
	\vspace{-0.2cm}
	\subsection{Framework}
	The framework of our method is described in Figure \ref{frame1}, where we adopt a pipeline method. In the upper stream, the best set of predicted entities is derived from optimal F1 threshold searching. In the downstream, input tokens with the predicted entities are sent to the entity-aware dialogue generation model. As a result, the final responses are obtained via the entity-revised diverse beam search.
	\vspace{-0.2cm}
	\subsection{Medical Entity Prediction}
		\begin{figure}[t]
		\centering
		\includegraphics[scale = 0.42]{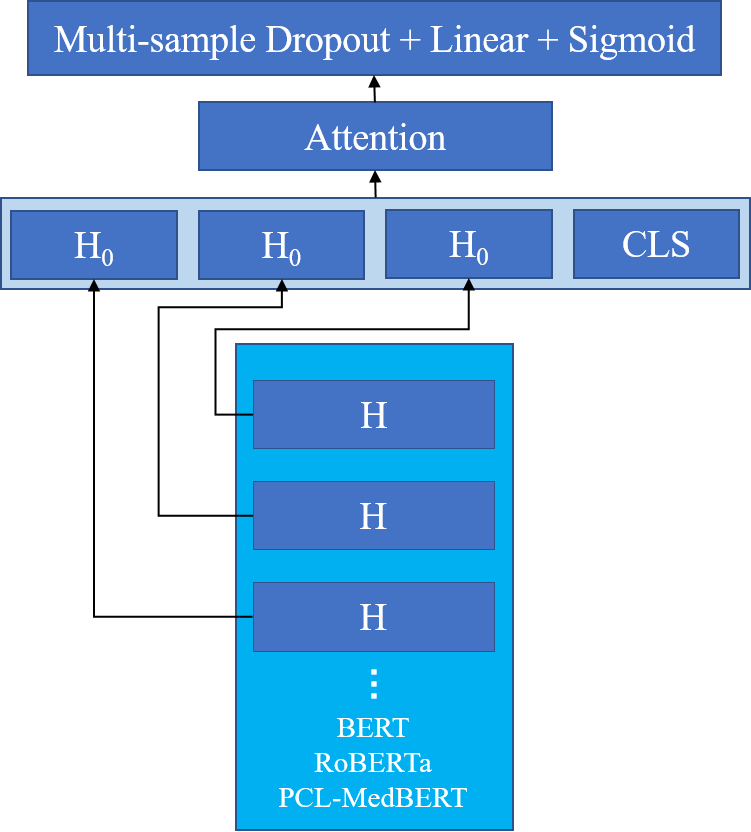}
		\caption{Structure of Our Entity Prediction Model.} \label{pred}
	\end{figure}
	Our medical entity prediction model is shown in Figure \ref{pred}, where we choose different pre-trained models, including BERT\cite{9}, RoBERTa\cite{10}, PCL-MedBERT\footnote[1]{https://code.ihub.org.cn/projects/1775}, RoBERTa-wwm-ext\cite{10.1} etc. as the backbone. As for the pre-trained models for general domains, we use the Mac-Bert's pre-training method\cite{11} with online medical data\footnote[2]{https://github.com/UCSD-AI4H/Medical-Dialogue-System \\ https://github.com/Toyhom/Chinese-medical-dialogue-data \\
	https://github.com/flyyuan/Chinese-Medical-QA-Data \\
	https://github.com/liuhuanyong/MiningZhiDaoQACorpus \\ 
	https://github.com/zhangsheng93/cMedQA2 \\
	https://github.com/lddsdu/VRBot
	} as continuing pre-training \cite{12}, in order to improve the generalization of the model in medical domain tasks.
	\par We extracted the features $H_{0}$ of the last three layers with the concatenation of the $CLS$ vector. Then, the concatenated vector is passed through an attention layer to utilize the information between different layers. The final predicted entity distribution is obtained via Multi-sample dropout\footnote[3]{https://github.com/lonePatient/multi-sample\_dropout\_pytorch}, fully perception layer, and sigmoid function. Specifically, we concatenate multiple rounds of dialogue history with history entities, and introduce $[SAP]$ token as a separator to separate history dialogues and history entities, such as $[CLS] + Sen1 +[SAP] + Ent1 +[SAP] + Ent1 + [SEP]+ Sen2 +[SAP] +Ent1^{\prime} + [SEP]$. As a result, the multi-category classification task is designed with the loss function $\mathcal{L}_{\mathrm{p}}(X, T)$, which is defined as follows:
	\begin{equation}
		\mathcal{L}_{\mathrm{p}}(X, T)=\frac{1}{N} \sum_{k=1}^{N}-w_{k}\left[t_{k} \cdot \log \sigma\left(x_{k}\right)+\left(1-t_{k}\right)-\log \left(1-\sigma\left(x_{k}\right)\right)\right]
	\end{equation}
	where $w_k$ is the optimal weight performed with F1 threshold search on the validation set, $t_k$ is the target entity and $x_k$ is the input feature.
	\vspace{-0.1cm}
	\subsection{Optimal F1 Threshold Search}
	As the categories are not balanced and the result obtained by cross-entropy loss is not globally optimal, in order to access the best F1 index, the optimal weight $w_k$ is designed with the optimal F1 threshold search. \par
	We consider each category of the multi-category problem as a two-category problem. A reasonable threshold can be obtained by threshold search. More precisely, we can obtain the most optimal threshold by adjusting the threshold from $0.3$ to $0.6$ through grid search, with the step of $0.001$.
	\vspace{-0.1cm}
	\subsection{Entity-aware Dialogue Generation}
	\begin{figure}[h]
		\centering
		\includegraphics[scale = 0.40]{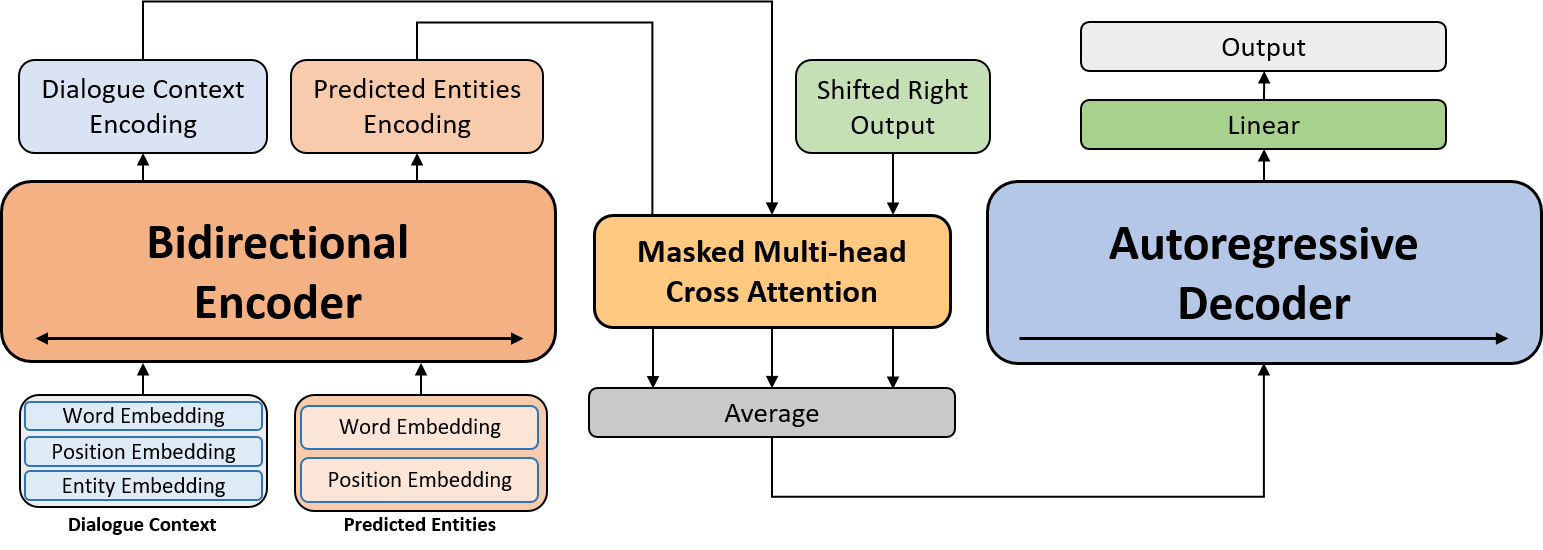}
		\caption{Entity-aware Dialogue Generation Model.} \label{EDG}
	\end{figure}
	The entity-aware dialogue generation model is presented in Figure \ref{EDG}. We adopt the encoder-decoder architecture as the backbone, where the dialogue context and the predicted entities are fused through the Masked Multi-head Cross Attention Mechanism (MMCA) \cite{6} , so that the predicted entity is used as the condition to generate the final response via auto-regression. The following part will introduce each component of the proposed model.
	\subsubsection{Context embedding module}
	\vspace{-0.2cm}
	\begin{figure}[h]
		\centering
		\includegraphics[scale = 0.4]{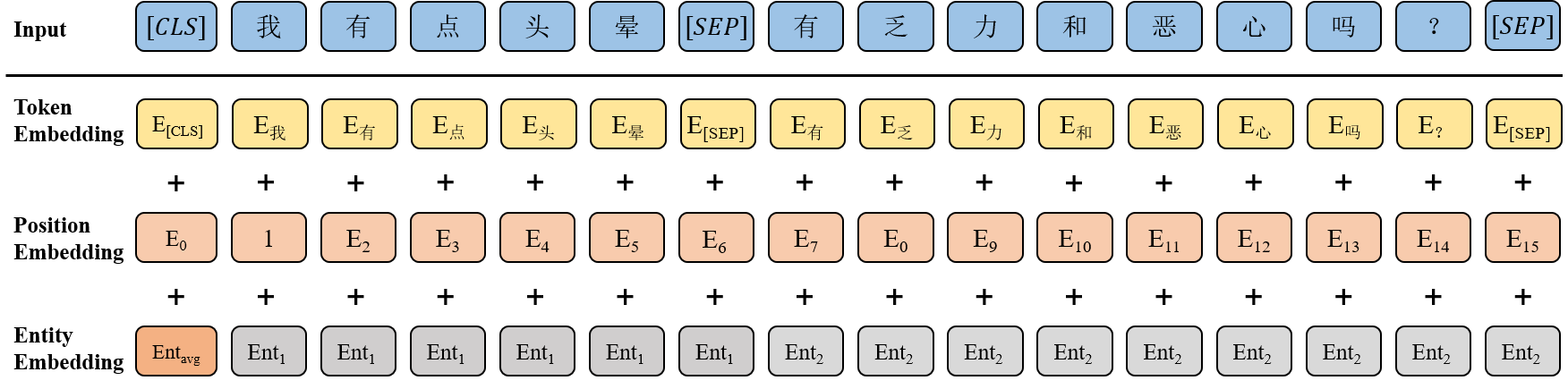}
		\caption{Structure of Dialogue Context Embedding.} \label{DCE}
	\end{figure}
	\vspace{-0.2cm}
	The context embedding module is presented in Figure \ref{DCE}, where the token embedding, the position embedding, and the entity embedding of the input are added together as the dialogue context embedding. Note that $E_{avg}$ is the average embedding of the entire sentence (divided by the sentence length), and the other part, such as $Ent_k$, is the embedding of the entity contained in the corresponding sentence $k$, which is obtained through a two-layer linear perception projection.
	\vspace{-0.2cm}
	\subsubsection{Predicted Entities embedding}
	The entities are concatenated together and separated by $[SEP]$, which is mapped in the form of tokens by the tokenizer. As a result, the entity embedding is obtained by adding these tokens with the position embedding.
	\vspace{-0.2cm}
	\subsubsection{Encoding Fusion Mechanism}
	We design the encoding fusion mechanism, where the encodings of dialogue context $E_C$, the predicted entities $E_{ent}$ and the shifted right output $E_{prev}$ are sent together to the Masked Multi-head Cross Attention module. The equations are shown as follows
	\begin{equation}
	O_{e n t}=\text{MMCA}\left[E_{\text {prev }}, E_{\text {ent }}, E_{\text {ent }}\right]
	\end{equation}
	 \begin{equation}
	 O_{C}=\text{MMCA}\left[E_{\text {prev }}, E_{\text {C }}, E_{\text {C }}\right]
	 \end{equation}
	where the $O_{ent}$ and the $O_{C}$ represent the encoding of predicted entities and dialogue context.
	In order to pay attention to the information of decoded tokens, the previous decoded encoding $O_{prev}$ is obtained through equation \ref{Oprev}
	 \begin{equation}
	 O_{prev}=\text{MMCA}\left[E_{\text {prev }}, E_{\text {prev }}, E_{\text {prev }}\right].
	 \label{Oprev}
	 \end{equation}
	Ater obtaining the $O_{ent}$, $O_{C}$ and $O_{prev}$, the averaging is performed, which is shown as equation \ref{avg}
	\begin{equation}
	O_{avg}=\left(O_{\text {ent}} + O_{\text {C}} + E_{\text {prev }}\right) / 3.
	\label{avg}
	\end{equation}
	\subsubsection{Dialogue Generation}
	The dialogue generation is processed via autoregressive decoding, the loss function is shown as follows
	\begin{equation}
	\begin{aligned}
	\mathcal L_{D}(\varphi) &=-\sum_{i} \log P_{\varphi}\left(y_{i} \mid x_{0}, \ldots, x_{i-1}, O_{C}, O_{ent}\right) \\
	&=-\sum_{i} \log P_{\varphi}\left(y_{i} \mid O_{avg}\right)
	\end{aligned}
	\end{equation}
	where $i$ represents the $i-th$ word generated by the decoder, and
	$x_0,... , x_{i-1}$, $y_i$ is a sequence of words from the generated response. Identically, the input of the decoder also can be represented as the mean fused encoding.
	\subsubsection{Auxiliary Tasks}
	Noted that there are two main gaps running in our framework, one is the gap between the data utilized in the	pre-training and fine-tuning stage, the other is the gap between the predicted entities and the real entities. As a result, we design two auxiliary tasks to fill the above-mentioned gaps
	\begin{enumerate}
	\item Language model task
	\begin{equation}
		\mathcal L_{L M}(\varphi)=-\sum_{i} \log P_{\varphi}\left(x_{i} \mid x_{i-k}, \ldots, x_{i-1}\right)
	\end{equation}
	where $\varphi$ representes the parameters of the encoder, $k$ is the size of the context window, and $x_{i-k}, \ldots, x_{i-1}$,$x_i$ is the sequence of tokens sampled from the training corpus.
	\item Hierarchical entity prediction task
	\begin{equation}
		\mathcal L_{T-5}(\theta)=-\sum_{i} \log P_{\theta}\left(t_{i} \mid O_{C}\right)
	\end{equation}
	\begin{equation}
		\mathcal L_{T-160}(\theta)=-\sum_{i} \log P_{\theta}\left(t_{i} \mid O_{C}\right)
	\end{equation}
	where $t_i$ is ground truth,  $\mathcal L_{T-5}(\varphi)$ represents the loss function of the 5 domain types, $\mathcal L_{T-160}(\varphi)$ represents the loss function of the 160 entity types.
	\end{enumerate}
	As a result, the final loss funciton can be written as
		\begin{equation}
	\mathcal L(\varphi, \theta)= \mathcal L_{D}(\varphi) + \mu \mathcal L_{L M}(\varphi) + \nu \mathcal L_{T-5}(\theta) + \lambda \mathcal L_{T-160}(\theta)
	\end{equation}
	where $\mathcal L(\varphi, \theta)$ is the total loss, and all the tasks share the same parameters. To facilitate training, we set the weights $\mu, \nu, \lambda$ to 1.
	\subsection{Entity-revised Diverse Beam Search}
	Diverse Beam Search is used to generate diversified responses. However, due to the lack of conditional information guidance, the results of the original DBS are often uncontrollable but diversified. Therefore, we design EDBS under predicted entity conditions during response generation, adopting the entity modification method as a guidance so that the results do not shift from the condition entities. Specifically, we considered the absence of true entities, the presence of incorrect entities, and the redundant predicted entities during the generation process. The algorithm design is as follows
	\begin{breakablealgorithm}
		\caption{Entity-revised Diverse Beam Search}
		\begin{algorithmic}[1]
			\renewcommand{\algorithmicrequire}{\textbf{Input:}} 
			\renewcommand{\algorithmicensure}{\textbf{Output:}}
			\Require
			Beam Search with a beam width of $B$;
			$\Omega \in (0,1)$;
			Converge threshold $\theta$; Edit distance length $L$; Output logits $\mathbf{y}$; Softmax function $\Theta$.
			\State{ \color{blue} // perform beam search without diversity for the first step} 
			\State {
				$Y_{[0]}^{1} \leftarrow \operatorname{argmax}_{\left(\mathbf{y}_{1,[0]}^{1} \ldots, \mathbf{y}_{B,[0]}^{1}\right)} \sum_{b \in\left[B\right]} \Theta\left(\mathbf{y}_{b,[0]}^{1}\right)$}
			\For {$t=1$ to $T$}
			\For {$k=1$ to $B$}
			\If {$\Omega \leq \theta$}
			\State{ \color{blue} // utilize multinomial to increase true entity sampling diversity} 
			\State{				$Y_{[t]}^{k} \leftarrow \operatorname{multinomial}_{\left(\mathbf{y}_{1,[t]}^{k} \ldots, \mathbf{y}_{B,[t]}^{k}\right)} \sum_{b \in\left[B\right]} \Theta\left(\mathbf{y}_{b,[t]}^{k}\right)$}
			\Else \State {
				$Y_{[t]}^{k} \leftarrow \operatorname{argmax}_{\left(\mathbf{y}_{1,[t]}^{k} \ldots, \mathbf{y}_{B,[t]}^{k}\right)} \sum_{b \in\left[B\right]} \Theta\left(\mathbf{y}_{b,[t]}^{k}\right)$}
			\EndIf
			\EndFor
			\State{ \color{blue} // ensure the final result can be converged } 
			\State {$\theta$ = $\theta$ * 0.9}
			\EndFor
			\State{ \color{blue} // perform the entity-revised method} 
			\For {$b$ in $B$}
			\State Divide each $b$ into list $G$ at sentence granularity 
			\State Mapping predicted entities $E$ into sentences list $S$
			\State Normalized the entities in each sentence $b$ to form the entity list $R$
			\For {$g$ in $G$}
			\For{$s$ in $S$}
			\If{Levenshtein\_Distance \footnote[4]{https://github.com/kylebgorman/EditTransducer} ($g$, $s$) $>$ $L$}
			\State{ \color{blue} // eliminate sentences containing redundant or wrong entities} 
			\State Delete the $l$ in $b$				
			\EndIf
			\EndFor
			\EndFor
			\For {$e$ in $E$}
			\If {$e$ not in $R$}
			\State{ \color{blue} // augment predicted entities into generated sentence} 
			\State Add entity $e$ into $b$
			\EndIf
			\EndFor
			\EndFor
			\Ensure
			Set of B solutions with the revised entities
		\end{algorithmic}
		\label{ccks1}
	\end{breakablealgorithm}
	To sum up, we consider the three cases of conditional shifting in DBS, and use the edit distance with entity condition to correct it until the final result is obtained.
	\vspace{-0.25cm}
	\subsection{Fusion Strategy}
	We adopt the curriculum 5-fold training strategy to fintune the Seq2Seq model, and use the bagging fusion mechanism to fuse the features of different models. As a result, two types of fusion mechanisms can improve the final performance. To a certain extent, These strategies can alleviate the problem of exposure bias, increasing quality of the final response.
	\vspace{-0.25cm}
	\section{Experiments}
	In this section, we will introduce the dataset, evaluation, implementation and results of the experiments.
	\subsection{Dataset}
	\vspace{-0.5cm} 
	\begin{table}[]
		\caption{Statistics of Dataset}
		\centering
		\setlength{\tabcolsep}{3mm}
		\renewcommand\arraystretch{1.2}
		\begin{tabular}{c|cccc}
			\noalign{\hrule height 1pt}
			& Train  & Dev(Test A) & Test(Ref) & Test B   \\ \hline
			\# Dialogues            & 17864  & 2747       & 452       & 1600   \\
			\# Utterance            & 385951 & 29601      & 4256      & 15750  \\
			\# Chars. per Dialogue  & 200.35 & 205.52     & 198.27    & 194.49 \\
			\# Chars. per Utterance & 18.9   & 19.07      & 19.8      & 19.75  \\
			\# Entity per Dialogue  & 6.33   & 7.99       & 8.24      & 8.13   \\
			\# Entity per Utterance & 0.79   & 0.74       & 0.82      & 0.83   \\ 
			\noalign{\hrule height 1pt}
		\end{tabular}
		\label{data}
	\end{table}
	\vspace{-0.5cm} 
	The dataset statistics are shown in Table \ref{data}. This task is based on a medical dialogue dataset with entity annotations. The Ref set is the test set sample announced online in advance.  Note that the number of entities and sentence length in the test set is more than the training set, which requires the model to be able to generate correct and longer sentences.
	\vspace{-0.3cm}
	\subsection{Evaluation}
	There are two indicators in the evaluation\footnote[5]{www.biendata.xyz/competition/ccks\_2021\_mdg/evaluation/}, including BLEU-avg\cite{13} score and Entity-F1. The BLEU-avg score is to measure response generation quality, while the Entity-F1 score is to measure entity correctness.
	
	\subsection{Implementation} 
	As for the entity prediction training method, we adopt the stratified learning rate with an attenuation strategy. Specifically, we set a larger value for the upper learning rate of the backbone, the internal learning rate of the pre-trained model is smaller, and the closer to the lower layer, the smaller the learning rate. We also adopt the FGM adversarial training\cite{14}, mixed-precision training \footnote[6]{https://github.com/NVIDIA/apex}, and moving average strategy to train the entity prediction model. \par
	For the dialogue generation part, we design curriculum boosting learning method to train the Seq2Seq model, which can be divided into three steps:
	\begin{enumerate}
 	\item The trained Seq2Seq model is utilized to initialize the parameters of the encoder and decoder, and fine-tune with the cleaning data. Finally, we use the boost method to train 4 epochs for a total of 5-fold;
	\item We use the dialogues with entities of all doctors for training, so that the generated response will contain the common features of doctors. We use the boost method to train 4 epochs for a total of 5-fold;
	\item We further sort out the dialogues with entities of doctors, whose length is greater than 11 (Counted on the Ref set) to train the model. As these dialogues have more entity characteristics. It is easier for the model to adapt to generating longer sentences. We train 2 epochs for a total of 5-fold.
	\end{enumerate}
	\vspace{-0.1cm} 
	\subsection{Result}
	In this Section, we will show our experimental results and online results.	
	\vspace{-0.4cm} 
	\subsubsection{Experimental Results}
		\begin{table}[t]
		\hspace{-1em}
		\begin{minipage}{\textwidth}
			\begin{minipage}[h]{0.45\textwidth}
				\centering
				\makeatletter\def\@captype{table}\makeatother\caption{F1 Performace in Backbone}			\renewcommand\arraystretch{1.2}	\setlength{\tabcolsep}{1mm}
				\begin{tabular}{cc} 
					\noalign{\hrule height 1pt}
					Model &  F1 \\
					\hline\noalign{\smallskip}
					BERT-base-chinese \cite{9} & 31.23 \\
					RoBERTa-wwm-ext \footnotemark[7] \cite{10.1}  & 31.68 \\
					RoBERTa-large \cite{10} &  33.23 \\
					Mac-BERT-large  \cite{11} & 33.64 \\
					PCL-BERT-wwm &   34.68 \\
					PCL-BERT-wwm-Post &  35.71 \\
					\noalign{\hrule height 1pt}	
	
				\end{tabular}
								\label{F1}
			\end{minipage}
			\begin{minipage}[h]{0.45\textwidth}
				\centering
				\makeatletter\def\@captype{table}\makeatother\caption{Final Results in Dev Data}		\renewcommand\arraystretch{1.2}	 \leftskip=-0.85em		\setlength{\tabcolsep}{1mm}
				\begin{tabular}{cccc}  
					\noalign{\hrule height 1pt}
					\vspace{-0.1cm}	
					Model &F1&Rec.&Acc.\\
					\noalign{\smallskip}\hline\noalign{\smallskip}	      
					RNN\_CNN &33.29&36.62&30.53 \\
					Last\_MaxPool&33.17&37.11&29.98\\
					Last3\_Embedding &34.43&38.22&31.32 \\
					Last3\_Attention &34.79&38.82&31.51 \\
					Last3\_MulDropout &35.30&37.74&33.16\\
					Last3\_Atten\_MulDrop &36.39&41.23&32.56 \\
					\noalign{\hrule height 1pt}

				\end{tabular}
				\label{Struct}
			\end{minipage}
		\end{minipage}
	\end{table}
	\footnotetext[7]{https://github.com/ymcui/Chinese-BERT-wwm}
	As is shown in the Table \ref{F1},	the symbol -Post represents continue pretraining with Mac-BERT pretraining method in the collected data. The BERT-base-chinese and RoBERTa-wwm-ext have similar effects. After the backbone being replaced by RoBERTa-large, the improvement is about 1.62. We finally choose the PCL-BERT-wwm as our baseline backbone, with the improvement of 1.03 after continue pretraining. We also try different model structures with the highest backbone as is shown in Table \ref{Struct}. The results show that it is competitive adopting the model structure with the concatenated features of the last three layers, attention mechanism and multi-dropout.

	\begin{table}[t]
		\centering
		\caption{Performance of Different Methods}
		\renewcommand\arraystretch{1.2}
		\setlength{\tabcolsep}{2mm}
		\begin{tabular}{lcccccc}
			\noalign{\hrule height 1pt}
			\vspace{-0.1cm} 
			& \multicolumn{3}{c}{Test A (Dev)} & \multicolumn{3}{c}{Test B}  \\
			\cmidrule(lr){2-4} \cmidrule(lr){5-7}
			Model                 & Avg. & F1 & BLEU & Avg. & F1 & BLEU \\ \hline
			Transformer \cite{15}          &     15.40        & 24.71   &  6.09       &  -           & -   & -         \\
			GPT2  \cite{16}       &     16.56        &  25.75  &    7.37      &   -          & -   &  -       \\
			BertGPT  \cite{6}            &      16.80       & 26.57   &    7.03    &   -         & -   &  -        \\
		 	\ \ \ + Curriculum 5-fold               &      19.42       & 28.72   &    10.12    &    19.16         & 28.34   &  9.99        \\
			T5-pegasus-small \cite{17}           &      16.55     & 23.76   &    9.34       &   -       & -  &  -       \\
			T5-pegasus-base  \cite{17}          &         17.42     & 25.41   &    9.43     &   -       &  -  &  -      \\
			T5-pegasus-base-Post  \footnotemark[8]   \cite{17}      &        17.58     & 25.55   &    9.61      &   -      &  -  &   -      \\
			CPM2-prompt \cite{18}  &       18.21  &  26.38  &    10.04     &  18.94           & 27.10   &  10.78      \\
			Logit-ensemble      &      19.83       &  28.92  &    10.74      &  20.73          & 29.54    & 11.92         \\	
      			\hline
			\ \ \ + Context Embedding      &      20.02       &  28.74  &    11.30      &  21.03          & 29.52    & 12.54         \\
			\ \ \ + Encoding Fusion     &      20.43       &  29.36  &    11.50      &  21.30          & 29.87    & 12.72         \\	
			\ \ \ + EDBS      &      21.24       &  30.12  &    12.36      &  21.83          & 30.57    & 13.09         \\	
			\noalign{\hrule height 1pt}
		\end{tabular}
	\label{Performance}
	\vspace{-0.3cm}
	\end{table}

\vspace{-0.3cm} 

	\begin{table}[]
		\caption{Performance of Different Decoding Methods}
		\centering
		\setlength{\tabcolsep}{2mm}
		\renewcommand\arraystretch{1.2}
		\begin{tabular}{lcccccc}
			\noalign{\hrule height 1pt}
			\vspace{-0.1cm} 
			& \multicolumn{3}{c}{Test A (Dev)} & \multicolumn{3}{c}{Test B}              \\
			\cmidrule(lr){2-4} \cmidrule(lr){5-7}
			Decoding   Method                   & Avg. & F1 & BLEU & Avg. & F1 & BLEU             \\ \hline
			Greedy                             &     20.43        &  29.36  & 11.50      &  21.30          & 29.87    & 12.72         \\
			Top\_k(k=20)                              &     20.23        &  29.03  &  11.43         &   -       & -  &  -                      \\
			Top\_p(p=0.9)   \cite{18}                             &     20.21        &  28.94  & 11.48         &   -       & -  &  -          \\		
			Top\_k\_p(k=20, p=0.9)  \cite{18}                             &     20.50        &  29.43  & 11.57          &   -       & -  &  -                     \\
			Multinomial   Sampling              &     20.53        &  29.23   &  11.83         &   21.35       & 29.90  &  12.80         \\			
			Beam Search  \cite{7}                       &    20.61         &  29.35  &  11.87          &  21.42       & 29.95  &  12.89         \\
			Diverse Beam Search    \cite{8}             &    20.76         &  29.58  &  11.95        &   21.60       & 30.24  &  12.95         \\
			\hline
			EDBS  &       21.24       &  30.12  &    12.36      &  21.83          & 30.57    & 13.09         \\
%
			\noalign{\hrule height 1pt}
		\end{tabular}
		\label{Decoded}
	\end{table}	

\vspace{-0.4cm} 
\footnotetext[8]{https://github.com/ZhuiyiTechnology/t5-pegasus}
	The results of different dialogue generation models are shown in the Table \ref{Performance}. The performance of the original Transformer\cite{15} is relatively poor, while the length of encoding limits the GPT2's \cite{16} ability (with small vocab). We carry out curriculum boost training for BertGPT\cite{6}, whose average score is 2.62 higher than the original one. With the scale of the pre-trained model increasing, each test score shows an upward trend. As a result, the fine-tuned CPM2-prompt\cite{18} reaches the highest score of 18.21 in average score among the single models. It can be found that the proposed context encoding module is beneficial to improve the BLEU score, as history entities are equally important for response generation. The encoding fusion module also improves  0.62 in F1 and 0.2 in BLEU effectively. We also compare different decoding methods with the proposed EDBS. As shown in Table \ref{Decoded}, it can be found that the EDBS method has significant advantages in entity accuracy and response quality compared to other methods.

		\begin{table}[t]
			\centering
			\renewcommand\arraystretch{1.2}
			\setlength{\tabcolsep}{4.155mm}
			\caption{Results of CCKS Competition}
			\begin{tabular}{ccccc}
				\noalign{\hrule height 1pt}
				\multicolumn{1}{c}{}     & \multicolumn{2}{c}{Test A}               & \multicolumn{2}{c}{Test B}                                      \\
				\cmidrule(lr){2-3} \cmidrule(lr){4-5}
				\multicolumn{1}{c}{Rank} & Team Name & \multicolumn{1}{c}{Score}    & Team Name                      & Score                          \\ \hline
				\multicolumn{1}{c}{1} & VPAI\_Lab & \multicolumn{1}{c}{21.24432} & { VPAI\_Lab} & {21.8283} \\
				\multicolumn{1}{c}{2}    & ParlAI    & \multicolumn{1}{c}{20.65078} & { sfeng}   & {21.4424} \\
				\multicolumn{1}{c}{3}    & sfeng     & \multicolumn{1}{c}{20.00431} & { iseesaw} & {21.0741} \\
				\multicolumn{1}{c}{4}    & iseesaw   & \multicolumn{1}{c}{19.03350}  & White\_jingling                & 20.0248                        \\
				\multicolumn{1}{c}{5}    & vivo      & \multicolumn{1}{c}{17.46282} & Little running snails                      &  19.0828                        \\ 
				\noalign{\hrule height 1pt}
			\end{tabular}
		\label{CCKS}
		\end{table}

		\begin{table}[t]
			\caption{Results of ICLR Workshop Competition}
			\vspace{-0.1cm} 
			\renewcommand\arraystretch{1.4}
			\scalebox{0.93}[1]{
			\begin{tabular}{ccccccccccc}
				
				\specialrule{1pt}{0pt}{0pt}
				\multicolumn{1}{c}{}     & \multicolumn{5}{c}{Phase A}                                  & \multicolumn{5}{c}{Phase B}            \\

				\cmidrule(lr){2-6} \cmidrule(lr){7-11}
				\multicolumn{1}{c}{Rank} & Team Name & Score & F1 & BLEU1 & \multicolumn{1}{c}{BLEU4} & Team Name & Score & F1 & BLEU1 & BLEU4 \\ \hline
				\multicolumn{1}{c}{1} &
				VPAI\_Lab &
				{ 32.43} &
				{ 23.67} &
				{ 47.90} &
				\multicolumn{1}{c}{{ 25.73}} &
				VPAI\_Lab &
				{ 36.03} &
				{ 29.58} &
				{ 50.48} &
				{ 28.03} \\
				\multicolumn{1}{c}{2} &
				jwanglvy &
				{ 32.09} &
				{ 18.07} &
				{ 49.66} &
				\multicolumn{1}{c}{{ 28.53}} &
				jackey &
				{ 35.28} &
				{ 31.72} &
				{ 49.09} &
				{ 25.04} \\
				\multicolumn{1}{c}{3} &
				zxxflyfish &
				{ 31.08} &
				{ 21.51} &
				{ 47.27} &
				\multicolumn{1}{c}{{ 24.46}} &
				LHS &
				{ 34.24} &
				{ 25.89} &
				{ 48.23} &
				{ 28.60} \\ 
			\noalign{\hrule height 1pt}
			\end{tabular}}
		\label{ICLR}
		\vspace{-0.4cm}
		\end{table}

\vspace{-0.4cm} 
\subsubsection{Online Results}
The detailed results are shown in Table \ref{CCKS} and Table \ref{ICLR}, our method wins both the CCKS and ICLR Workshop MLPCP Track 1 competitions. These results demonstrate that the proposed method is effective and solid.
\vspace{-0.7cm} 
\section{Conclusion}
\vspace{-0.2cm} 
	In this paper, we propose a pipeline framework for Chinese medical dialogue generation, which consists of two parts: medical entity prediction, and entity-aware dialogue generation. In our framework, 
	we first optimize the entity prediction model with F1 threshold search, then utilize the predicted entities with the proposed encoding fusion mechanism, which controls the information from different sources.
	The entity prediction model with F1 threshold search is used at the upstream. The predicted entities is ultilized with the proposed encoding fusion mechanism in the downstream, which controls the information from different sources.
	We improve the original DBS with the entity-revised method, which proves to be effective for bettering the quality of the final response.
	We win the best results in the CCKS and the ICLR Workshop MLPCP Track 1 competitions, which demonstrates the effectiveness and practicality of our proposed method.
	In the future, we will consider using the knowledge graph to infer the predicted entity, and try different fusion strategies when generating, to further improve the correctness and quality of the generated response.
	%
	%
	%
	%
	
\end{document}